\documentclass[twoside,leqno,twocolumn]{article}

\usepackage[letterpaper]{geometry}

\usepackage{xcolor}
\usepackage{ltexpprt}
\usepackage{hyperref}
\usepackage{subcaption}
\usepackage{multirow}
\usepackage{xspace}
\usepackage{amsmath}
\usepackage{graphicx}
\usepackage{balance}

\begin{document}

\newcommand{\stitle}[1]{\noindent{\textbf{#1}}}
\newcommand{\ititle}[1]{\vspace{1ex}\noindent{\em #1}}
\newcommand{\hadis}[1]{\textcolor{red}{(hadis) #1}}
\newcommand{\sara}[1]{\textcolor{green}{(sara) #1}}
\newcommand{\simone}[1]{\textcolor{violet}{(simone) #1}}
\newcommand{\new}[1]{\textcolor{black}{ #1}}
\newcommand{\blue}[1]{\textcolor{blue}{ #1}}
\newcommand{\red}[1]{\textcolor{red}{ #1}}
\newcommand{\tocite}{\red{\cite{}}\xspace}

\newcommand\relatedversion{}

\title{Fairness and Bias in Truth Discovery Algorithms:\\ An Experimental Analysis}
% \author{Corey Gray\thanks{Society for Industrial and Applied Mathematics.}
% \and Tricia Manning\thanks{Society for Industrial and Applied Mathematics.}}
%\author{Simone Lazier, Saravanan Thirmuruganathan, Hadis Anahideh \thanks{\{slazie2, hadis\}@uic.edu; University of Illinois Chicago.} }
%Sara: updated the author block based on the Siam template.. single author tag is frowned up on by ACM DL. 
\author{
Simone Lazier\thanks{University of Illinois Chicago. (slazie2@uic.edu)}
\and Saravanan Thirmuruganathan\thanks{QCRI, HBKU. (sthirumuruganathan@hbku.edu.qa)}
\and Hadis Anahideh\thanks{University of Illinois Chicago. (hadis@uic.edu)}
}
\date{}

\maketitle

% Copyright Statement
% When submitting your final paper to a SIAM proceedings, it is requested that you include
% the appropriate copyright in the footer of the paper.  The copyright added should be
% consistent with the copyright selected on the copyright form submitted with the paper.
% Please note that "20XX" should be changed to the year of the meeting.

% Default Copyright Statement
\fancyfoot[R]{\scriptsize{Copyright \textcopyright\ 2023 by SIAM\\
Unauthorized reproduction of this article is prohibited}}

% Depending on which copyright you agree to when you sign the copyright form, the copyright
% can be changed to one of the following after commenting out the default copyright statement
% above.

%\fancyfoot[R]{\scriptsize{Copyright \textcopyright\ 20XX\\
%Copyright for this paper is retained by authors}}

%\fancyfoot[R]{\scriptsize{Copyright \textcopyright\ 20XX\\
%Copyright retained by principal author's organization}}

%\pagenumbering{arabic}
%\setcounter{page}{1}%Leave this line commented out.

\begin{abstract}
Machine learning (ML) based approaches are increasingly being used 
in a number of applications with societal impact. 
Training ML models often require vast amounts of \emph{labeled} data, 
and crowdsourcing is a dominant paradigm for obtaining labels from multiple workers. 
%A large number of popular datasets in domains as diverse as 
%computer vision, natural language processing, and speech processing
%leverage crowdsourcing for labels.
Crowd workers may sometimes provide unreliable labels, and to address this, truth discovery (TD) algorithms such as majority voting are applied to determine the consensus labels from conflicting worker responses. However, it is important to note that these consensus labels may still be biased based on sensitive attributes such as gender, race, or political affiliation.
% Since crowd workers can provide inaccurate labels,
% truth discovery algorithms such as majority voting are used 
% to compute the consensus labels from conflicting worker responses. 
% However, such labels could be biased 
% based on sensitive attributes such as gender, race, and other attributes such as political affiliation. 
Even when sensitive attributes are not involved,
the labels can be biased due to different perspectives of subjective aspects such as toxicity.
In this paper, we conduct a systematic study of 
the bias and fairness of TD algorithms.
Our findings \new{using two existing crowd-labeled datasets}, reveal that a non-trivial proportion of workers provide biased results, 
and using simple approaches for TD is sub-optimal. 
Our study also demonstrates that popular TD algorithms are not a panacea. 
 Additionally, we quantify the impact of these unfair workers on downstream ML tasks 
and show that conventional methods for achieving fairness and correcting label biases are ineffective in this setting. 
% show that common approaches for achieving fairness and correcting label biases do not work well in this setting. 
We end the paper with a plea for the design of novel bias-aware truth discovery algorithms
that can ameliorate these issues.   
\end{abstract}
\section{Introduction}
\label{sec:intro}

Machine learning (ML) is becoming increasingly pervasive 
and is already having a significant impact on people's lives. 
Often, the decisions by these models impact different sub-groups in a non-uniform manner. 
Hence, there has been a flurry of research in designing fairness-aware machine learning algorithms. 
In this paper, we argue that such care must also be accorded to a crucial step in the ML process 
-- collection of data labels. 
We situate our work at the intersection of two major paradigms -- data-centric AI and fair ML. 
The former seeks to build AI systems with good-quality data 
while the latter ensures that the resulting system is fair. 
As we shall show shortly, the current process of label collection is simplistic
and results in biased labels that have a disproportionate impact 
on the accuracy and fairness of the downstream ML models. 
While there has been extensive work on designing fair ML algorithms,
there is limited work on developing algorithms to generate correct (and fair) labels.  

\stitle{Making of Data Sausage.}
Datasets are indispensable to the progress of the ML community 
allowing the design of novel models and serving as a benchmark for measuring advancements. 
Despite their importance, the vast majority of the work in ML 
is geared towards the development of new algorithms rather than data stewardship practice \cite{peng2021mitigating}. 
Recently, the fairness community has unearthed many issues with prominent benchmark datasets,
 such as label errors and bias \cite{northcutt2021pervasive,prabhu2020large}. 
In this work, we systematically analyze another source of bias -- 
labels generated through the simple aggregation of responses from biased crowd workers.% labels produced by simplistic aggregation of the responses produced by biased crowd workers. 
% \hadis{do we need to differentiate the bias definition in our context? 
%  lets check the paper and fix any inconsistent cases}

A key factor in the success of ML is the availability of large \emph{labeled} datasets.
However, creating accurate labels using a domain expert is prohibitively expensive.
Hence, a common alternative is to use crowd workers for generating the labels.
The workers are often hired using crowdsourcing marketplaces such as Amazon Mechanical Turk. 
Typically, the quality of these workers is sub-par compared to that of domain experts 
-- albeit significantly more cost-effective.
% at a much-reduced cost.
Quality control is often maintained by collecting multiple labels for each data point 
from various workers and implementing additional filtering mechanisms, such as pre-tests \new{where a set of `golden tasks' with known correct answers are given to workers before they begin labeling the main dataset \cite{zheng2017truth}.}
It is possible that the labels of the workers conflict with one another.
In this case, an aggregation method, more generally Truth Discovery algorithms (TD),
is used to determine the final label.
One common approach is to use majority voting. 
In many cases, the labels thus obtained are reasonably accurate and can be used to train ML models. 
A large number of major benchmark datasets were obtained using this approach.
Even considering a single domain such as Computer Vision, 
a non-exhaustive list of such datasets include
\new{ImageNet \cite{deng2009imagenet}, Places \cite{zhou2017places}, MS COCO {\cite{lin2014microsoft}}, ActivityNet {\cite{caba2015activitynet}}, Open
Street Map (OSM) dataset {\cite{haklay2008openstreetmap}}, YouTube-8M {\cite{abu2016youtube}}, and Intrinsic Images
in the Wild {\cite{bell2014intrinsic}}}.
% , and Materials in Context Database {I couldn't find it}.
The prevalence of crowdsourced datasets in other ML fields is equally significant.
For example, as much as 10\% of the papers published in Natural Language Processing 
uses crowdsourced dataset \cite{shmueli2021beyond}.

The pervasiveness of using crowdsourcing for obtaining labels is problematic.
The use of human annotators is a double-edged sword.
While humans can provide answers to challenging and subjective questions,
their responses are also potentially subject to a variety of biases.
With ML increasingly being utilized in sensitive applications, such as the detection of fake news, toxicity, and hate speech, the labels for these domains can be particularly subjective and involve nuanced concepts.
% ML is increasingly being used in more sensitive applications 
% including the detection of fake news, toxicity, and hate speech.
% For these domains, labels can be particularly subjective as they involve nuanced concepts. 
Different workers may assign different labels to the same data point 
based on their lived-in experiences, which can result in the injection of subjective values and biases into the dataset. 
% In other words, the subjective values and biases of the annotators could potentially get injected into the dataset. 
Using datasets with extensive amounts of biased labels can lead to models that are  sub-par in terms of accuracy and fairness.

\stitle{Summary of Our Contributions.}
We conduct a systematic investigation of the aggregation process for labels in crowdsourced datasets.
% process by which the labels are aggregated for crowdsourced datasets.
% We conducted our experiments over 
Our study includes two \new{publicly available} diverse datasets -- \emph{Crowd Judgement} and \emph{Twitter Toxicity}, and the results were consistent with those from our experiments on a number of other datasets in hate speech detection. 
% Our experiments over a number of other datasets in hate speech detection show similar findings. 
Our findings indicate that biased %(and unfair) \hadis{we never differentiated between the two} 
% \simone{I'd focus only on unfair, biased sounds confusing to me} \hadis{now checking other papers and reading the rest of the paper I think bias refers to the worker's implicit inclination and unfairness refers to the modeling outcome. But maybe we should clarify and be consistent. later we use unfair workers. I am not sure yet let's discuss} 
workers are widely prevalent 
and cause a significant impact on the quality of the dataset labels. 
Surprisingly, it is possible for a worker to be accurate while also being biased \new{, even though the dataset with the true labels might be biased itself}.
Hence, this precludes simplistic solutions such as removing biased workers 
as it could have collateral damage on the accuracy of the other items. 
There has been extensive work on truth discovery algorithms 
that seek to aggregate conflicting labels into a single `truth'.
However, we find that using such TD algorithms, which are more complex than majority voting, is not effective in solving the problem. 
We conduct analyses using two popular algorithms (DS and LFC) \new{, Dawid Skene (DS) \cite{dawidskene1979} and Learning From Crowds (LFC) \cite{raykar2010learning}, } that have different probabilistic generative modeling for crowd worker bias. %\hadis{which model are we referring to? here I think the use of bias is confusing}.
Nevertheless, their performance is not substantially better than that of majority voting. 
Furthermore, we study the impact of the biased labels on downstream ML tasks 
and show that each of the truth discovery algorithms suffers from similar flaws,
leading to less accurate and unfair models.  
We also show that using fair ML algorithms does not eliminate the issue of biased labels.

We make the following key contributions:
% we make the following contributions. 
\begin{enumerate}
    \item We identify truth discovery algorithms for aggregating the responses of crowd workers as a major source of bias in crowdsourced datasets.
    \item We conduct an extensive experimental study to examine the various facets of this bias and quantify the impact on downstream ML tasks.
    \item We offer practical recommendations for practitioners %provide guidance for practitioners 
    and highlight promising research problems. 
\end{enumerate}

\stitle{Paper Organization.}
The rest of the paper is organized as follows. 
In Section \ref{sec:background}, we present the relevant background 
on crowdsourcing, fairness, and truth discovery algorithms. Section \ref{sec:methadology} outlines the datasets and research methodology used in our study.
Our experimental analysis is structured along five main dimensions, described in detail in Sections \ref{sec:h1}-\ref{sec:h5}.
%We summarize the major findings along with guidance for practitioners in Section \ref{sec:discussion}. 
We conclude with some parting thoughts in Section \ref{sec:conclusion}.

\section{Background}\label{sec:background}

\stitle{Crowdsourcing for ML and its Pitfalls.}
Over the last decade, crowdsourcing has been extensively used  to advance machine learning research.
Crowd workers have been used for data generation, label generation, and evaluation of ML models. 
The ML community has embraced crowdsourcing as a way to obtain image and linguistic annotations 
needed for training popular vision and language processing tasks. 
A key stumbling block is that the labels provided by the workers can be noisy or inaccurate. 
Workers can be imperfect, unmotivated, or influenced by various cognitive biases \cite{eickhoff2018cognitive}. 
When the labels are generated for training ML models for tasks with societal impact,
these biases can have serious consequences.
A pioneering work identified various sources of bias that can affect ML algorithms \cite{mehrabi2021survey}.
Truth discovery algorithms have been proposed to address the former issues 
while fair ML algorithms have been proposed to address the latter. 

\stitle{Truth Discovery for Crowdsourcing.}
Achieving quality control in crowdsourcing involves determining the correct answers 
from the low-quality answers provided by crowd workers.
Unfortunately, the vast majority of works use simple approaches 
such as majority voting, which is often insufficient.
Sophisticated approaches are often based on probabilistic generative models 
that allow the modeling of various factors on the worker label generation.
These factors could be worker-based (accuracy, bias, motivation, collusion, topic-related expertise, etc.) or 
task-related (difficulty, clarity, etc.).
A taxonomy of truth discovery algorithms can be found in \cite{zheng2017truth}.

\stitle{Truth Discovery - Problem Setup.}
Consider a set of tasks $\mathcal{T} = \{t_1, \ldots, t_n\}$ with the size of $n$. 
Let $\mathcal{W}$ be the set of annotators.  
and $W_i$ be the vector of answers for task $t_i$ collected by a subset of annotators in $\mathcal{W}$. 
Let $m$ be the total number of annotators. 
Each task $t_i$ could be labeled by a subset of workers. 
By querying from these annotators, 
we would like to determine the true class label $y_i$ for any task $t_i \in \mathcal{T}$. 
We then denote the label of annotator $j$ for task $t_i$ as $l_{ij}$. 
Let $\tilde{y}_i^w$ be the answer of annotator $w$ for task $t_i$ and $y$ be the unknown ground truth. 
Given the answers matrix of $\tilde{Y}$, 
the ground truth inference of $y_i$ for each task is identified as $\hat{y}_i$ using a truth discovery algorithm.

\stitle{Algorithmic Fairness in ML.}
In recent years, the study of algorithmic fairness 
has gained increasing attention in the machine learning community. 
The objective of these studies is to ensure that 
AI systems do not discriminate against certain groups of individuals 
based on sensitive features such as race, gender, or age. 
A detailed description of the various aspects can be found in \cite{barocas-hardt-narayanan}. 
Despite the growing body of literature on fairness in machine learning, 
there are still a number of challenges and limitations that need to be addressed. 
For example, there is currently no widely accepted definition of fairness, 
which can lead to inconsistencies in the assessment of fairness in AI systems \cite{mehrabi2021survey}.
Additionally, many existing methods for mitigating algorithmic bias 
rely on adjustments to the modeling processes 
including pre-processing \cite{zhang2018mitigating}, 
in-processing \cite{agarwal2018reductions}, and post-processing \cite{lohia2019bias} phases, 
rather than addressing the root causes of bias (e.g., data collection bias), 
which is especially the case in crowdsourcing settings.
% \hadis{do we need to refer to the KDD paper here? and there is another paper Hindawi satisfies fairness and privacy}
\section{Experimental Methodology}
\label{sec:methadology}

In this paper, we are interested a specific type of bias that results in unfair labels.
For example, workers could provide incorrect labels for tasks involving members of a specific gender or race.
Using TD algorithms such as majority voting perpetuates these biased labels into the dataset
which in turn results in unfair models.
Ideally, these unfair labels should be corrected during the label aggregation process. 

In this section, we outline the experimental methodology we use to systematically study and 
quantify how truth discovery algorithms work in the presence of unfair workers. %the fairness and bias \hadis{do we analyze both?} of truth discovery algorithms. 
For ease of exposition, we have partitioned them into five research hypotheses. 
After describing the hypotheses, we provide more details about the experimental setting, 
such as the list of representative truth discovery algorithms and datasets. 

\stitle{Experimental Hypotheses.}
We conduct an extensive set of experiments to study 
how different TD algorithms impact the aggregation of labels from imperfect workers.
Specifically, we focus on five key dimensions 
that expound the issue of bias in label aggregation.
Sections \ref{sec:h1}-\ref{sec:h5} describe the corresponding research hypotheses in detail.
In this section, we provide a high-level overview
so that the reader has a holistic perspective of our experiments.

\ititle{H1: \new{Quantifying Prevalence of Unfair Workers through Analysis of Accuracy and Fairness Metrics.}}
We begin by conducting an exploratory analysis of the 
accuracy and fairness metrics of individual workers. 
The results indicate that unfair workers are not anomalies and are instead widely prevalent. 
% \simone{I'd rephrase this last sentence, maybe something like "We show that, depending on a chosen threshold, a significant portion of workers can be considered unfair."} \hadis{well I do not agree on bc then we need to explain the threshold here it is just a general plan}

\ititle{H2: \new{Impact of Unfair Workers on Label Flipping in Truth Discovery Algorithms.}
}
% Next, we investigate the impact of unfair workers on label aggregation.
% Specifically, we focus on majority voting which is widely used.
% \simone{Maybe since before we say that "Unfortunately, the vast majority of works use simple approaches such as majority voting which is often insufficient." this would need to be justified} \hadis{Simone you can add a sentence here to clarify}
% \hadis{well I think all td algs are insufficient if ignore the bias. here we just want to show the impact of biased workers not TDs unfairness}
We show that unfair workers can potentially flip the labels for a non-trivial proportion of tasks. 
{Next, we investigate the impact of unfair workers on TD algorithms to show that unfair workers can potentially flip the labels for a non-trivial proportion of tasks. Specifically, we focus on majority voting, which is a widely used method. It is worth noting that this issue affects not only majority voting but also other traditional truth discovery algorithms.}
% , but every traditional TD algorithm is affected by this issue. 
% -- Like This?}

\ititle{H3: \new{Unfair Workers Reduce the Performance of TD Algorithms.}}
Next, we investigate the performance of three representative truth discovery algorithms.
Specifically, we compare the ground truth labels estimated by these methods with the true ground truth.
We show that all of these algorithms have comparable (and sub-optimal) accuracy and fairness.
This highlights the need for the development of new fairness-aware truth discovery algorithms, rather than relying on existing ones.
% This allows us to conclude that it is desirable to 
% design novel fairness-aware truth discovery algorithms instead of reusing existing truth discovery algorithms. 

\ititle{H4: \new{Unfair Workers Negatively Impact the Downstream ML Tasks.}}
The next hypothesis addresses a potential concern that 
the study of the performance of truth discovery algorithms is not merely academic.
Instead, we demonstrate that training an ML model with the output of 
simple truth discovery algorithms result in a model that is both less accurate and less fair. 

\ititle{H5: \new{The Complementarity of Fair ML and Fair TD.}}
In the final hypothesis, 
we explore the idea that using fair ML algorithms cannot completely eliminate the bias introduced by
traditional truth discovery algorithms. {We also evaluate the effectiveness of FairTD, a fair truth discovery algorithm proposed in \cite{kdd2020}}. 
% \hadis{we also have results of fairTD}

\stitle{Truth Discovery Algorithms.}
There has been extensive work on techniques for aggregating conflicting responses from workers. 
In this paper, we compare three representative algorithms.
For a detailed discussion of their comparative performance,
please refer to \cite{zheng2017truth}. 
\emph{Majority Voting (MV)} is a simple and widely used algorithm 
that chooses the response provided by the majority of the workers as the consensus.
MV might work if the workers are reasonably competent and do not form any collusion.
A key limitation of MV is that it weighs the response of each worker equally. 
However, this might be sub-optimal in a crowd setting where workers can have divergent accuracy.
For example, if only one worker is an expert while others are novices, MV may fail.
In such cases, it is important to model the worker accuracy 
so as to compute a \emph{weighted} majority voting. 
Dawid-Skene (DS) \cite{dawidskene1979} is a more sophisticated model 
that assigns different weights to workers based on the content of their confusion matrix. 
Worker accuracy is estimated using an iterative EM algorithm. 
Learning from Crowd (LFC) \cite{raykar2010learning} is a probabilistic model 
that proposes a two-coin approach for annotators.
Each worker has two biased coins with bias corresponding to sensitivity and specificity.
The worker flips one of these coins based on the true label of the task 
and uses the outcome to determine their response.
This approach has been used successfully in many domains, such as 
aggregating the predictions of radiologists. 
% Each of these three models has been widely used and has had empirical success. 
All three models have been widely used and have shown empirical success. 
In this paper, we investigate how these models perform in the presence of worker bias.

\stitle{Fairness Metrics.}
There has been extensive work on measuring and mitigating bias in datasets and ML models. 
However, there is not much consensus on the appropriate fairness metrics 
as different metrics require that some aspect of the ML model remains comparable to all subgroups. %\hadis{I could not find any discussion on adopting the notions for crowd}
% \hadis{I feel like these are more ml-based definitions should we revise them for the crowd setting? I need to come back to this part}
In this paper, we use two key widely used  metrics that are designed to evaluate the fairness of ML models -- 
demographic parity and equalized odds. 
By treating individual crowd workers as fallible oracles (similar to ML models),
we can reuse these two metrics for measuring the fairness of workers. 
Demographic parity (DP) is achieved when the probability of a particular prediction (such as loan disbursal)
is not dependent on sensitive group membership (such as gender=Male).
Equalized odds (EO) fairness metric requires that the predictions of the ML model be 
both independent of sensitive group membership 
\emph{and} have comparable true and false positive rates. 
% seeks to ensure that an ML model 
% performs equally well for different subgroups. 
It is important to note that EO is stricter than DP.
% as it requires that the predictions of the ML model be 
% both independent of sensitive group membership 
% \emph{and} have comparable true and false positive rates. 
% Specifically, we use two derived metrics for these two fairness metrics --  difference and ratio.
To evaluate the fairness of our models more effectively, we use two derived metrics for DP and EO -- difference and ratio. 
A difference of 0 indicates the DP/EO has been achieved, and a ratio of 1 indicates the same. 
These derived metrics are useful in that they allow us to express the fairness metrics as a single scalar for easier comparison.
% to convert the fairness metrics into a single scalar for comparison.  

\stitle{Datasets.}
\new{We conduct our experiments using two existing diverse and representative datasets from different domains.
The \emph{Crowd Judgement} dataset \cite{dressel2018accuracy} contains labels of the workers from a major crowdsourcing marketplace
for 1000 cases from the infamous COMPAS dataset for recidivism.} 
Groups of 20 workers evaluated the same set of 50 defendants. 
% The workers had no expertise in criminal justice, and 
% their responses had accuracy and bias comparable to that of the original COMPAS model. 
\new{The second dataset \emph{Jigsaw Toxicity} contains labels of workers on whether a Twitter comment is toxic. This dataset has been collected in \cite{JigsawToxicityDatasetRef}}
We used a subset of 14000 comments that were labeled for toxicity, obscenity, threats, insults, and hate. 
% The workers for these tasks were more qualified and provided with appropriate metrics.

\section{H1: \new{Quantifying Prevalence of Unfair Workers through Analysis of Accuracy and Fairness Metrics}}
\label{sec:h1}

\begin{figure*}
  \begin{subfigure}[t]{0.24\textwidth}
    \includegraphics[width =\textwidth]{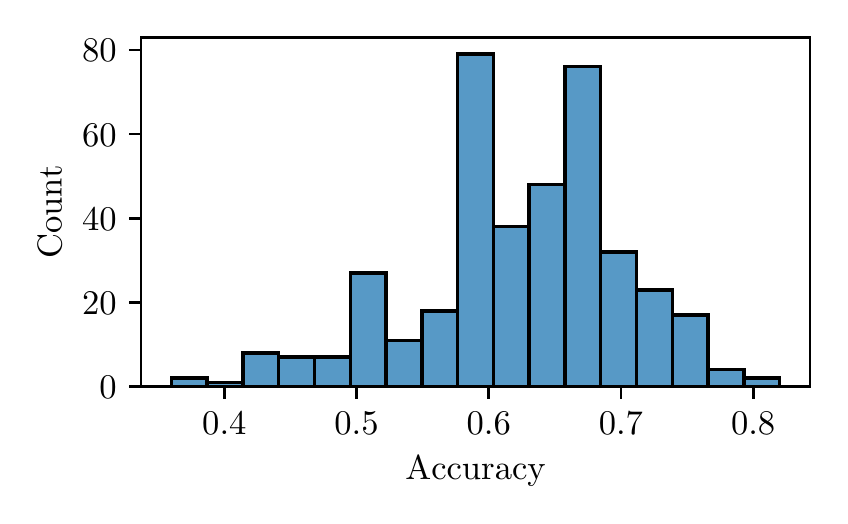}
    \caption{Accuracy}
    \label{fig:h1_cj_accuracy}
  \end{subfigure}\hfill
    \begin{subfigure}[t]{0.24\textwidth}
      \includegraphics[width =\textwidth]{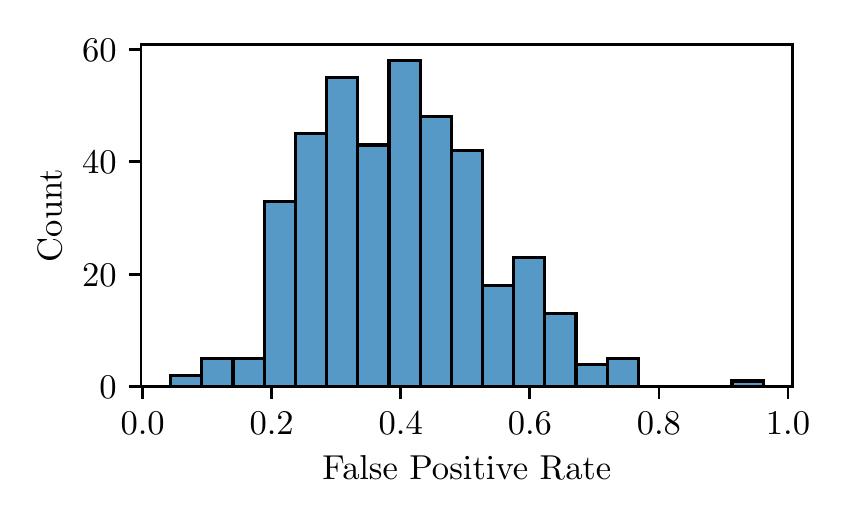}
      \caption{False Positive Rate}
      \label{fig:h1_cj_false_positive_rate}
  \end{subfigure}\hfill
  \begin{subfigure}[t]{0.24\textwidth}
    \includegraphics[width =\textwidth]{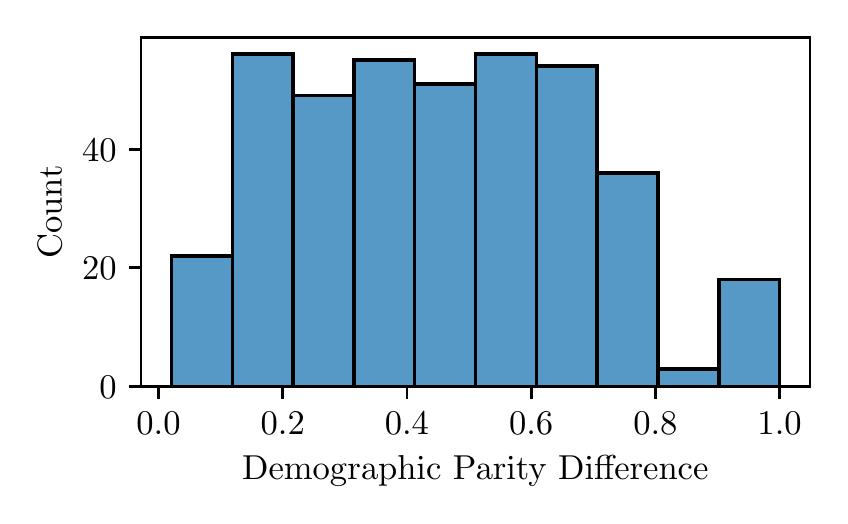}
    \caption{Demographic Parity }
    \label{fig:h1_cj_demographic_parity_difference}
  \end{subfigure}
  \begin{subfigure}[t]{0.24\textwidth}
    \includegraphics[width =\textwidth]{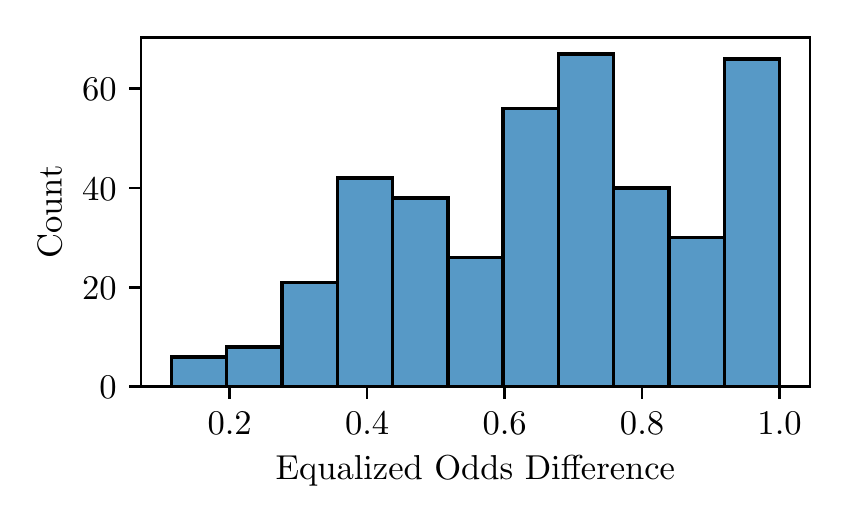}
    \caption{Equalized Odds }
    \label{fig:h1_cj_equalized_odds_difference}
  \end{subfigure}\hfill
  \caption{Histogram of Accuracy and Fairness Metrics of Workers of \emph{Crowd Judgement} Dataset}
  \label{fig:h1_cj_accuracy_fairness_metrics}
\end{figure*}

\begin{figure*}
    \begin{subfigure}[t]{0.24\textwidth}
      \includegraphics[width =\textwidth]{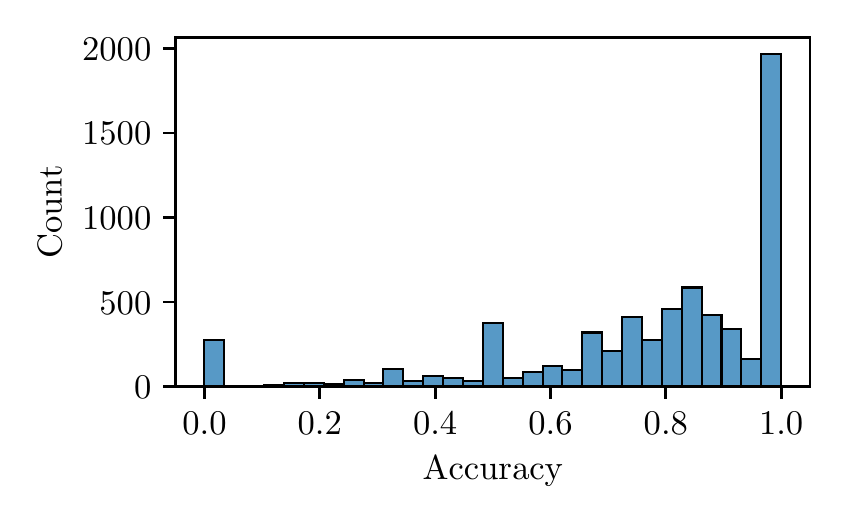}
      \caption{Accuracy}
      \label{fig:h1_jt_accuracy}
    \end{subfigure}\hfill
    \begin{subfigure}[t]{0.24\textwidth}
      \includegraphics[width =\textwidth]{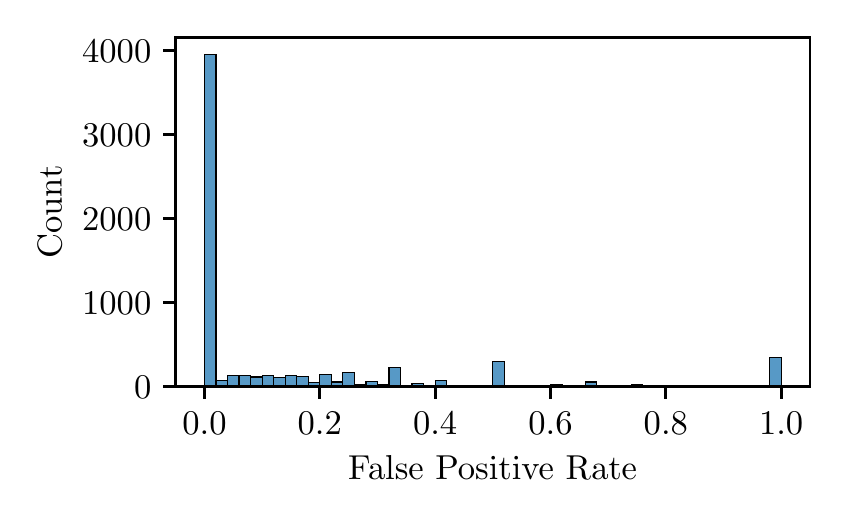}
      \caption{False Positive Rate}
      \label{fig:h1_jt_false_positive_rate}
    \end{subfigure}\hfill
    \begin{subfigure}[t]{0.24\textwidth}
      \includegraphics[width =\textwidth]{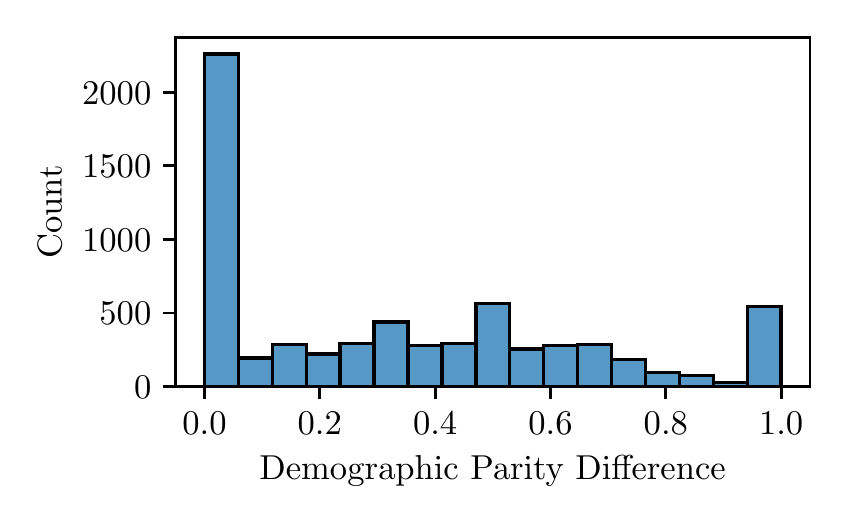}
      \caption{Demographic Parity }
      \label{fig:h1_jt_demographic_parity_difference}
    \end{subfigure}
    \begin{subfigure}[t]{0.24\textwidth}
      \includegraphics[width =\textwidth]{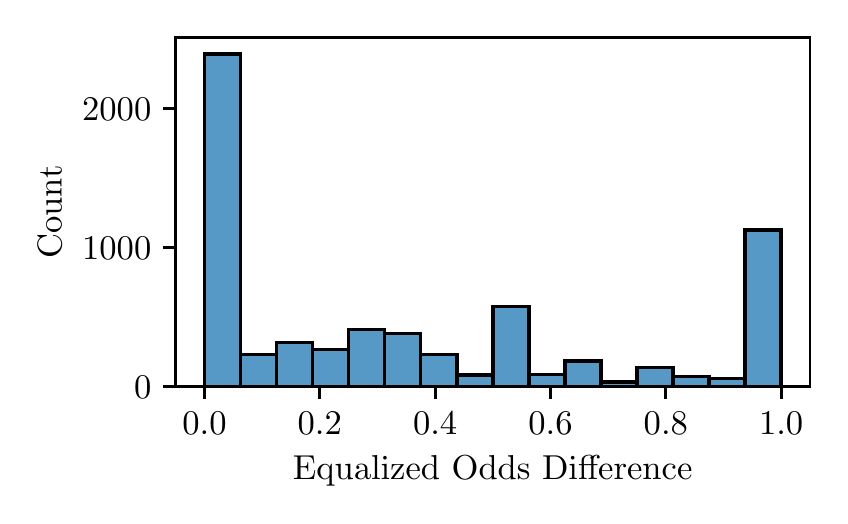}
      \caption{Equalized Odds }
      \label{fig:h1_jt_equalized_odds_difference}
    \end{subfigure}\hfill
    \caption{Histogram of Accuracy and Fairness Metrics of Workers of \emph{Jigsaw Toxicity} Dataset}
    \label{fig:h1_jt_accuracy_fairness_metrics}
  \end{figure*}

Our first set of experiments conducts an exploratory analysis of the dataset to compute the accuracy and fairness metrics of the workers.
Despite conventional wisdom that workers are often imperfect and/or unfair, quantifying these factors is crucial for understanding the challenges truth discovery algorithms face.
% it is often important to precisely quantify them.
% This allows us to have a better handle on the challenges 
% that must be overcome by truth discovery algorithms. 

Both the datasets (\emph{Crowd Judgement} and \emph{Jigsaw Toxicity}) consist of ground truth.
This contrasts with typical crowdsourcing datasets where the ground truth is often unavailable.
This metadata allows us to measure worker accuracy and fairness. 
Specifically, we focus on two accuracy-related metrics (accuracy and false positive rate)
and two fairness-related metrics (demographic and equalized odds difference).
We conduct three interrelated sets of experiments. 
In the first set of experiments, 
we generate histograms of the four accuracy and fairness metrics.
Second, we create a scatter plot that contrasts an accuracy metric with a fairness metric.
Finally, we show the accuracy-fairness distribution in an alternative tabular format
as it provides unique insights.

\stitle{Distribution of Worker Accuracy Metrics.}
Figures \ref{fig:h1_cj_accuracy} and \ref{fig:h1_cj_false_positive_rate}
display the histogram of worker accuracy and false positive rates for the \emph{Crowd Judgement} dataset. 
We can see that worker accuracy ranges approximately between 0.4 and 0.8.
In other words, the workers are neither perfect (with an accuracy close to 1.0)
nor incompetent (with an accuracy close to 0.0).
Nevertheless, a non-trivial number of workers have a substantial false positive rate, underscoring the challenges truth discovery algorithms face.
% In general, this shows the challenges faced by truth discovery algorithms.
The workers are very accurate and often suffer from high false positive rates. 
Hence, using simple aggregation metrics such as majority voting is unlikely to be sufficient. 

The corresponding histogram for \emph{Jigsaw Toxicity} datasets is
in Figures \ref{fig:h1_jt_accuracy} and \ref{fig:h1_jt_false_positive_rate}.
The worker accuracy distribution distinct from that of the \emph{Crowd Judgement} dataset.
For this dataset, most workers are highly accurate, with a considerable number having perfect accuracy, and few exhibit a high false positive rate. 
% For this dataset, the workers are reasonably accurate, with many workers having perfect accuracy.
% Similarly, most workers have meager false positive rates. 
Even though the two datasets have very different workers in terms of accuracy,
we shall show that both contain a non-trivial amount of unfair workers. 

\stitle{Distribution of Worker Fairness Metrics.}
Figures \ref{fig:h1_cj_demographic_parity_difference} and \ref{fig:h1_cj_equalized_odds_difference}
shows the histogram of the demographic parity and equalized odds difference metrics 
of the workers for the \emph{Crowd Judgement} dataset, respectively. 
Recall that a value closer to 0 indicates a fairer worker.
The histograms of both these metrics show that workers exhibit a wide divergence in fairness.
While there are some fair workers, the vast majority are unfair to some degree. 
The histograms for the \emph{Jigsaw Toxicity} dataset, presented in Figures \ref{fig:h1_jt_demographic_parity_difference} and \ref{fig:h1_jt_equalized_odds_difference}, show that the workers are generally fairer than those of the \emph{Crowd Judgement} dataset.
% A similar observation can be made for the Jigsaw Toxicity dataset 
% whose histograms can be found in 
% Figures \ref{fig:h1_jt_demographic_parity_difference} and \ref{fig:h1_jt_equalized_odds_difference}.
% We can see that the workers of this dataset are fairer than that of the Crowd Judgement dataset.
This is unsurprising due to the \new{golden tasks} and clearer rubrics. 
% \hadis{can you explain?}
Nevertheless, contain a significant number of unfair workers. 
\new{We can also conclude that simply adding more workers will not solve this problem as a significant portion 
of these workers will be unfair too. 
Hence, it is imperative to design a truth discovery algorithm that cognizant of the large proportion of unfair workers for better aggregating worker labels. 
}

\begin{figure*}[!htb]
    \centering
    \begin{minipage}{.49\textwidth}
        \begin{subfigure}[t]{0.49\textwidth}
            \includegraphics[width =\textwidth]{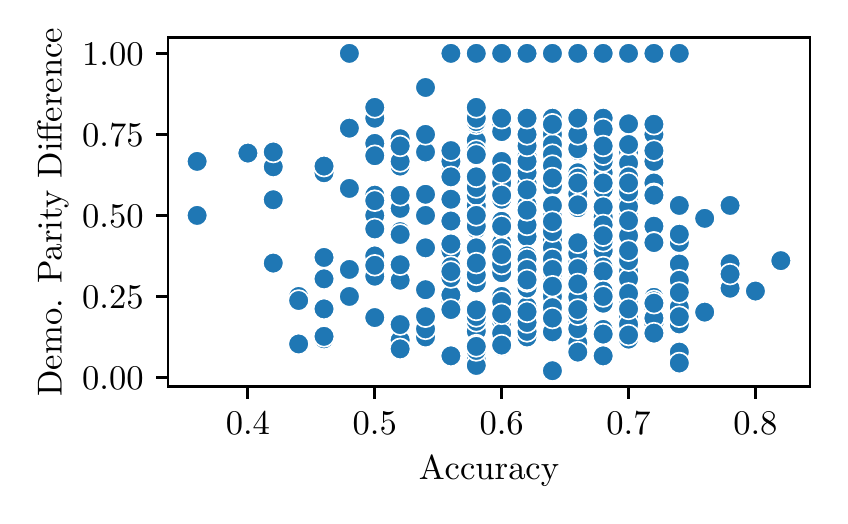}
            \caption{Accuracy vs Demographic Parity Difference}
            \label{fig:h1_cj_acc_vs_demo_par_diff}
          \end{subfigure}\hfill
          \begin{subfigure}[t]{0.49\textwidth}
            \includegraphics[width =\textwidth]{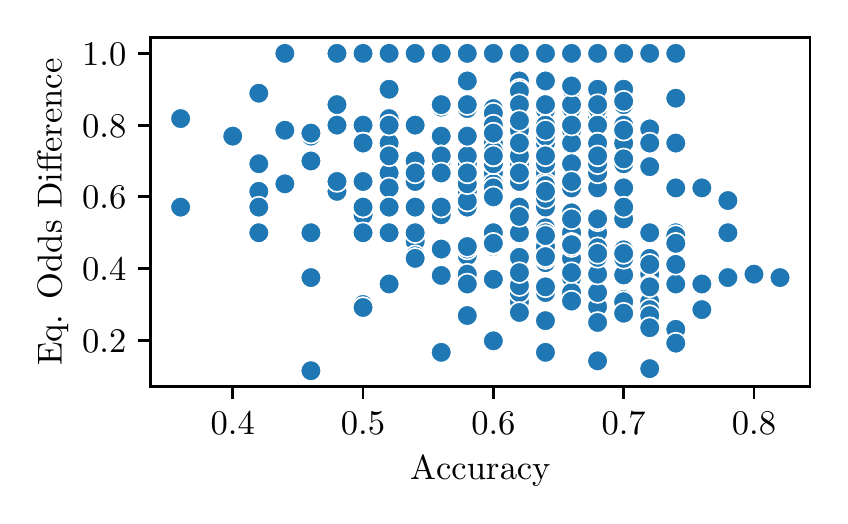}
            \caption{Accuracy vs Equalized Odds Difference}
            \label{fig:h1_cj_acc_vs_eq_odds_diff}
          \end{subfigure}
          \caption{\emph{Crowd Judgement} dataset}
          \label{fig:cj_acc}
    \end{minipage}%
    \hfill
    \begin{minipage}{0.49\textwidth}
        \begin{subfigure}[t]{0.49\textwidth}
            \includegraphics[width =\textwidth]{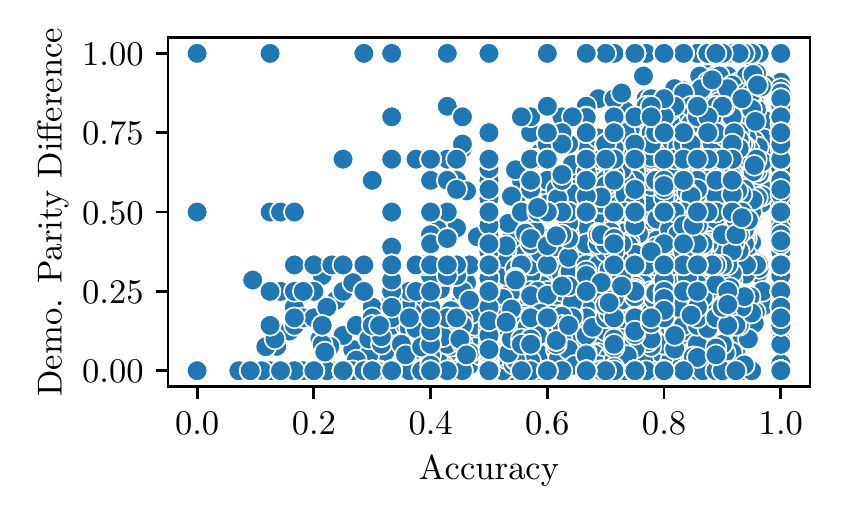}
            \caption{Accuracy vs Demographic Parity Difference}
            \label{fig:h1_jt_acc_vs_demo_par_diff}
          \end{subfigure}\hfill
          \begin{subfigure}[t]{0.49\textwidth}
            \includegraphics[width =\textwidth]{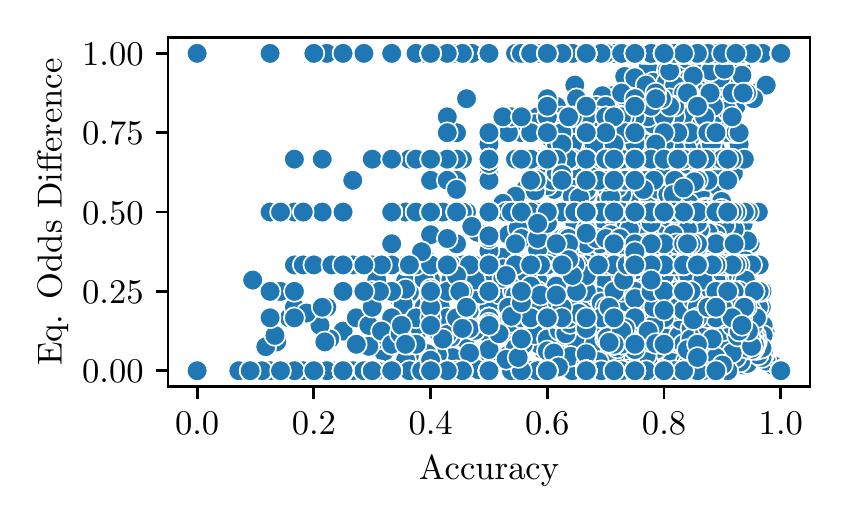}
            \caption{Accuracy vs Equalized Odds Difference}
            \label{fig:h1_jt_acc_vs_eq_odds_diff}
          \end{subfigure}
          \caption{\emph{Twitter Toxicity} dataset}
          \label{fig:jt_acc}
    \end{minipage}
\end{figure*}

\stitle{Worker Accuracy-Fairness Trade-off.}
A natural question that might arise is about the correlation between inaccurate and unfair workers.
In other words, are inaccurate workers also unfair?
If there is a high correlation between the metrics, 
then developing a separate class of fairness-aware truth discovery algorithms is unnecessary.
We can study the relationship between these metrics by 
plotting an accuracy metric against a fairness metric. 
Figure \ref{fig:h1_cj_acc_vs_demo_par_diff} show the scatter plot
of worker accuracy on the X-axis and 
worker demographic parity difference on the Y-axis.
Figure \ref{fig:h1_cj_acc_vs_eq_odds_diff} shows a similar plot for 
accuracy and equalized odds difference metrics.
We can see values spread over the entire plot 
without tight clusters showing relationships between accuracy and fairness. 
In other words, a worker can be both accurate and unfair.
The corresponding plots for the \emph{Jigsaw toxicity} dataset in Figures \ref{fig:h1_cj_acc_vs_demo_par_diff} and \ref{fig:h1_cj_acc_vs_eq_odds_diff} shows that the number of unfair workers increases as the accuracy increases. Thus, workers can be both accurate and unfair, and it is insufficient to directly use traditional truth discovery algorithms that are geared towards accuracy in the presence of unfair workers. 

Table \ref{tbl:accuracyVsFairnessMetrics} 
shows a granular version of Figures \ref{fig:cj_acc} and \ref{fig:jt_acc}
by first bucketing workers based on their accuracy 
and then computing the fairness metrics for all workers in this bucket.
Once again, we see no strong correlation between the two metrics.
For the \emph{Crowd Judgement} dataset, there is a weak correlation 
where the more accurate the worker, the lower their unfairness (i.e., the metric trends towards 0).
However, this effect is absent for the toxicity dataset.

\begin{table*}[!htb]
  \centering
  \scriptsize
    \begin{tabular}{|c|c|c|c|c|}
    \hline
                            & \multicolumn{2}{c|}{\textbf{Crowd Judgement}}                            & \multicolumn{2}{c|}{\textbf{Twitter Toxicity}}                           \\ \hline 
    \textbf{Accuracy Range} & \multicolumn{1}{l|}{\textbf{Demo. Par. Diff}} & \textbf{Eq. Odds. Diff} & \multicolumn{1}{l|}{\textbf{Demo. Par. Diff}} & \textbf{Eq. Odds. Diff} \\ \hline
    (0.0, 0.1{]}            & \multicolumn{1}{l|}{NA}                       & NA                      & \multicolumn{1}{l|}{0.05}                     & 0.05                    \\ 
    (0.1, 0.2{]}            & \multicolumn{1}{l|}{NA}                       & NA                      & \multicolumn{1}{l|}{0.12}                     & 0.15                    \\ 
    (0.2, 0.3{]}            & \multicolumn{1}{l|}{NA}                       & NA                      & \multicolumn{1}{l|}{0.11}                     & 0.22                    \\ 
    (0.3,  0.4{]}           & \multicolumn{1}{l|}{0.62}                     & 0.72                    & \multicolumn{1}{l|}{0.24}                     & 0.35                    \\ 
    (0.4, 0.5{]}            & \multicolumn{1}{l|}{0.49}                     & 0.67                    & \multicolumn{1}{l|}{0.26}                     & 0.42                    \\ 
    (0.5, 0.6{]}            & \multicolumn{1}{l|}{0.45}                     & 0.69                    & \multicolumn{1}{l|}{0.3}                      & 0.5                     \\ 
    (0.6, 0.7{]}            & \multicolumn{1}{l|}{0.46}                     & 0.67                    & \multicolumn{1}{l|}{0.34}                     & 0.53                    \\ 
    (0.7, 0.8{]}            & \multicolumn{1}{l|}{0.4}                      & 0.55                    & \multicolumn{1}{l|}{0.37}                     & 0.5                     \\ 
    (0.8, 0.9{]}            & \multicolumn{1}{l|}{0.36}                     & 0.38                    & \multicolumn{1}{l|}{0.45}                     & 0.43                    \\ 
    (0.9, 1.0{]}            & \multicolumn{1}{l|}{NA}                       & NA                      & \multicolumn{1}{l|}{0.32}                     & 0.22                    \\ \hline
    \end{tabular}
    \caption{Distribution of Accuracy and Fairness metrics for both datasets.}
    \label{tbl:accuracyVsFairnessMetrics}
\end{table*}

%%%%%%%%%%%%%%Sara: commenting this out as this is duplicate of the other plots that I am showing
%%%%% violin plot is over convoluted and does not add any information over simple histogram.
%\begin{figure}[!htb]
%\begin{subcaptionblock}{0.45\linewidth}
%\includegraphics[width=\linewidth]{plots/Accuracy vs eq odds - compas.pdf}
%\vspace{-2.5cm}
%\caption{Equalized Odds vs Accuracy (COMPAS)}
%\end{subcaptionblock}
%\hfill
%\begin{subcaptionblock}{0.45\linewidth}
%\vspace{0.5cm}
%\includegraphics[width=\linewidth]{plots/Equalized odds distribution - compas.pdf}
%\vspace{-2.5cm}
%\caption{Equalized Odds (COMPAS)}
%\end{subcaptionblock}
%\caption{Worker statistics for COMPAS}
%\label{fig:h1compas}
%\end{figure}

%\begin{figure}[!htb]
%\begin{subcaptionblock}{0.45\linewidth}
%\includegraphics[width=\linewidth]{plots/Accuracy vs eq odds - twitter.png}
%\caption{Equalized Odds vs Accuracy (Twitter)}
%\end{subcaptionblock}
%\hfill
%\begin{subcaptionblock}{0.45\linewidth}
%\vspace{0.5cm}
%\includegraphics[width=\linewidth]{plots/Equalized odds distribution - twitter.png}
%\caption{Equalized Odds (Twitter)}
%\end{subcaptionblock}
%\caption{Worker statistics for Twitter toxicity}
%\label{fig:h1twitter}
%\end{figure}
\section{H2: \new{Impact of Unfair Workers on Label Flipping in Truth Discovery Algorithms}}
\label{sec:h2}

In the next set of experiments, we investigate how the 
presence of unfair workers impacts the labels produced by the MV algorithm, widely used for aggregating crowd worker labels in most ML datasets.
We approach this investigation from two different perspectives.

\begin{figure*}[!htb]
    \centering
    \begin{minipage}{.49\textwidth}
      \begin{subfigure}[t]{0.49\textwidth}
        \includegraphics[width =\textwidth]{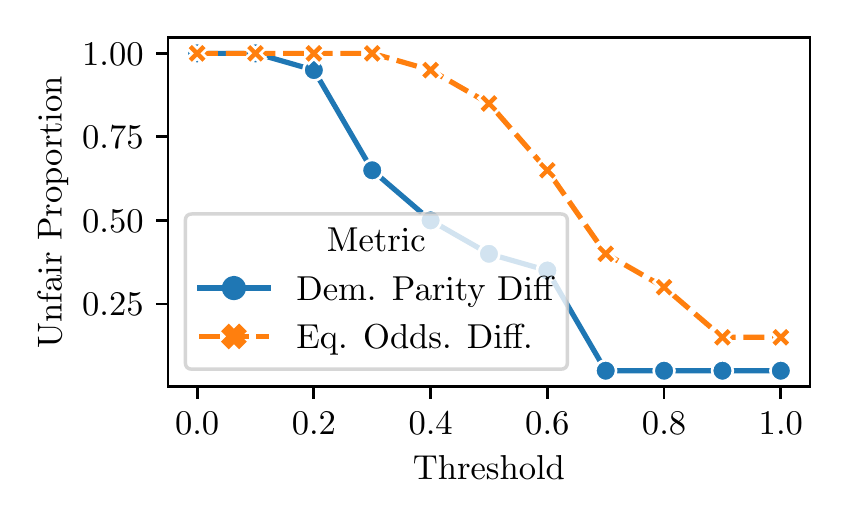}
        \caption{\emph{Crowd Judgement}}
        \label{fig:h2_cj_threshold_unfair_prop}
      \end{subfigure}
      \begin{subfigure}[t]{0.49\textwidth}
        \includegraphics[width =\textwidth]{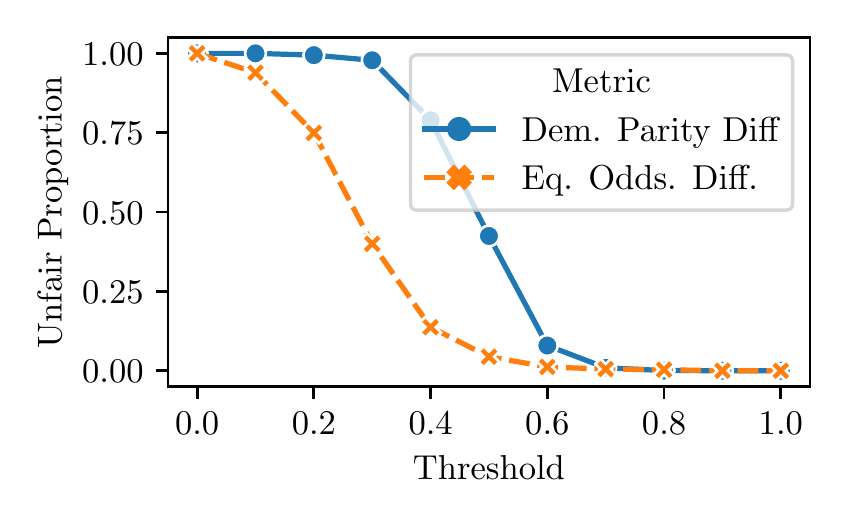}
        \caption{Jigsaw Toxicity}
        \label{fig:h2_jt_threshold_unfair_prop}
      \end{subfigure}
          \caption{Proportion of tasks dominated by unfair workers for different thresholds of fairness}
          \label{fig:h2_unfair_prop}
    \end{minipage}%
    \hfill
    \begin{minipage}{0.49\textwidth}
      \begin{subfigure}[t]{0.49\textwidth}
        \includegraphics[width =\textwidth]{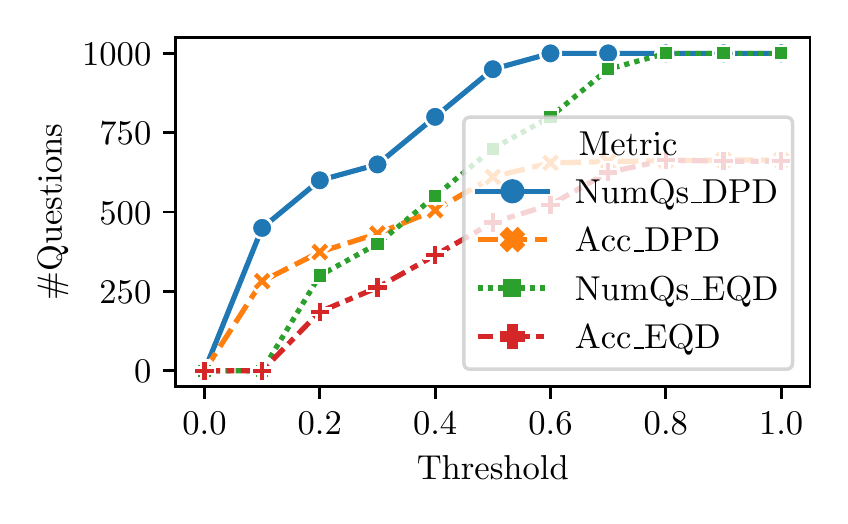}
        \caption{\emph{Crowd Judgement}}
        \label{fig:h2_cj_threshold_accuracy_impact}
      \end{subfigure}\hfill
      \begin{subfigure}[t]{0.49\textwidth}
            \includegraphics[width =\textwidth]{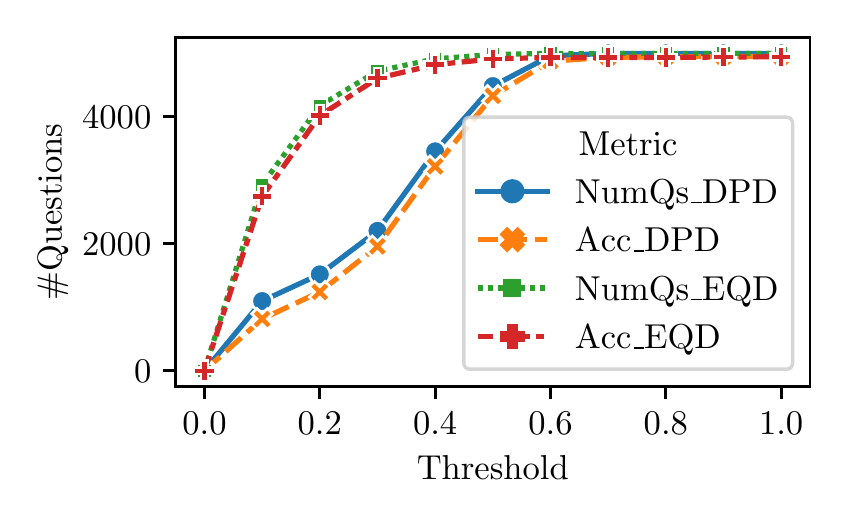}
            \caption{\emph{Jigsaw Toxicity}}
            \label{fig:h2_jt_threshold_accuracy_impact}
      \end{subfigure}\hfill
          \caption{Impact of removing unfair workers on overall accuracy}
          \label{fig:h2_threshold_accuracy_impact}
    \end{minipage}
\end{figure*}

\stitle{Proportion of Unfair Workers for Tasks.}
Intuitively, MV chooses the response 
provided by the majority of the workers as the consensus.
Hence, if the majority of the workers assigned to a task turns out to be unfair, 
then the output of the task will turn out to be unfair. 
Our first set of experiments investigates how the proportion of workers assigned to a task changes based on the definition of fairness. We use the demographic parity and equalized odds difference metrics to determine the fairness of a worker, where a value of 0.0 indicates a lack of bias. To designate a worker as fair or unfair, we apply a threshold to these fairness metrics.
Clearly, a worker with a score of 0.0 is fair while someone with a score of 1.0 is unfair. 
We systematically vary the minimum threshold for fairness (from $0, 0.1, \ldots 1.0$ ).
% Based on the threshold, a worker will be considered fair or unfair.
We then count the proportion of tasks where unfair workers 
outnumber fair workers thereby poisoning the output of majority voting.
Figure \ref{fig:h2_unfair_prop} plots how the proportion varies 
for both datasets using two fairness metrics. 
We can see that even with a relaxed threshold of 0.2,
%(a common threshold widely used such as in the four-fifths rule),
a large proportion of tasks are dominated by unfair workers. 
% \hadis{I think it still needs clarification. for threshold 0.8, zero percent of the tasks are dominated by unfair workers, is not it?}
This observation once again shows that it is important to design fairness-aware TD algorithms. 

\stitle{Impact of Removing Unfair Workers.}
A natural rejoinder might be to eliminate unfair workers 
and run TD algorithms on the remainder.
In our next set of experiments, we show the futility of this approach. 
Recall from Figures \ref{fig:cj_acc} and \ref{fig:jt_acc}
that it is possible for a worker to be broadly accurate while being unfair. 
Figure \ref{fig:h2_threshold_accuracy_impact} shows the impact of removing unfair workers. 
% \hadis{removing unfair workers?}
When we remove workers designated as unfair based on the fairness metrics used in the previous experiment, it can lead to changes in two key parameters.
Firstly, removing unfair workers could potentially affect the accuracy of the tasks. This is measured using the Acc\_DPD and Acc\_EQD  metrics for demographic parity and equalized odds, respectively.
Secondly, the number of tasks with responses could also change.
Recall from Figure \ref{fig:h2_unfair_prop} that unfair workers dominate in many tasks. 
So it is plausible that \emph{all} workers are unfair for some tasks, leading to the abandonment of the task. 
% In this case, we have to potentially abandon this task.
The metrics NumQs\_DPD and NumQs\_EQD measure this phenomenon.
From Figure \ref{fig:h2_threshold_accuracy_impact},
we can see that the overall accuracy drops when we remove unfair workers.
This is due to the fact that a good chunk of them are accurate 
and removing them results in the collateral damage of reduced accuracy.
Moreover, as we drop unfair workers,
the number of abandoned tasks increases. \new{This fact suggests that removing unfair workers is not practical for such tasks.}
From both these observations, we can conclude that 
simply dropping unfair workers is a sub-optimal solution
as it results in lower accuracy. 
% \hadis{do we drop both inaccurate and unfair?}
To overcome this challenge, it is crucial to design fairness-aware TD algorithms that can counteract unfair workers without compromising accuracy.

\new{A key research challenge is that truth discovery algorithms are \emph{unsupervised} in nature. Note that if the ground truth for the crowdsourcing tasks were known, workers would not have been needed. These algorithms often take an EM style iterative approach. They begin with an estimate (such as random initialization) of the relevant metric such as worker accuracy and improve the estimate in each iteration. While the approach for improving estimates of worker accuracy is well known, designing the corresponding algorithms for fairness is an open research problem.}

\section{H3: \new{Unfair Workers Reduce the Performance of TD Algorithms.}}
\label{sec:h3}

In this section, we study how different TD algorithms perform in the presence of unfair workers.
Recall that these algorithms are often designed for workers who are inaccurate.
We conduct our experiment as follows.
We run three representative TD algorithms over the worker responses to obtain an approximate estimate of the ground truth labels.
Since both \emph{Crowd Judgment} and \emph{Jigsaw Toxicity} provide us with actual ground truth, 
we can compare this with the output of the TD algorithms.
This also allows us to compute the accuracy and fairness metrics for each of these TD algorithms.
Tables \ref{tbl:tdAlgorithmsCrowdJudgement} and \ref{tbl:tdAlgorithmsJigsawToxicity}
show the result of this experiment.

At a high level, we can see that the accuracy of the three algorithms is roughly comparable,
with the more complex TD algorithms based on generative modeling (DS and LFC) 
providing slightly more accurate estimates. 
We observe a similar pattern in the fairness metrics, with none of the TD algorithms able to effectively handle the unfair workers. Using advanced TD algorithms seems to exacerbate this issue, as some  underlying assumptions, such as the two-coin annotator process for LFC, are violated by the presence of unfair workers. 
% This is not surprising as the presence of unfair workers violates
% the assumptions of the generative process of the TD algorithm 
% (such as the two-coin annotator process for LFC).
% \simone{Should we say which assumptions?} \hadis{an example is provided for LFC} 
We can conclude that it is sub-optimal to directly use existing TD algorithms, and instead, it is necessary to design fairness-aware TD algorithms 
that explicitly model the unfairness of the workers. 

\begin{table*}[]
\scriptsize
    \begin{tabular}{|l|c|c|c|c|c|}
    \hline
    \textbf{TD Algorithm} & \textbf{Accuracy} & \textbf{\begin{tabular}[c]{@{}l@{}}Demographic \\ Parity Difference\end{tabular}} & \textbf{\begin{tabular}[c]{@{}l@{}}Demographic Parity \\ Ratio\end{tabular}} & \textbf{\begin{tabular}[c]{@{}l@{}}Equalized Odds \\ Difference\end{tabular}} & \textbf{\begin{tabular}[c]{@{}l@{}}Equalized Odds \\ Ratio\end{tabular}} \\  \hline
    Majority Voting       & 0.658             & 0.166                                  & 0.706                             & 0.12                               & 0.703                         \\ 
    Dawid-Skene           & 0.663             & 0.183                                  & 0.667                             & 0.136                              & 0.665                         \\ 
    Learning from Crowd   & 0.664             & 0.184                                  & 0.668                            & 0.142                              & 0.676                         \\ \hline 
    \end{tabular}
    \caption{Performance of different TD algorithms for \emph{Crowd Judgement} dataset}
    \label{tbl:tdAlgorithmsCrowdJudgement}
\end{table*}

% \red{claim for the following table JT uses detoxify as baseline and not MV}

\begin{table*}[]
\scriptsize
    \begin{tabular}{|l|c|c|c|c|c|}
    \hline
    \textbf{TD Algorithm} & \textbf{Accuracy} & \textbf{\begin{tabular}[c]{@{}l@{}}Demographic \\ Parity Difference\end{tabular}} & \textbf{\begin{tabular}[c]{@{}l@{}}Demographic Parity \\ Ratio\end{tabular}} & \textbf{\begin{tabular}[c]{@{}l@{}}Equalized Odds \\ Difference\end{tabular}} & \textbf{\begin{tabular}[c]{@{}l@{}}Equalized Odds \\ Ratio\end{tabular}} \\  \hline
    Majority Voting       & 0.92              & 0.607                                                                             & 0.08                                                                                         & 0.016                                                                         & 0.08                                                                     \\ 
    Dawid-Skene           & 0.93              & 0.636                                                                             & 0.08                                                                                         & 0.051                                                                         & 0.05                                                                     \\ 
    Learning from Crowd   & 0.95              & 0.64                                                                              & 0.08                                                                                         & 0.06                                                                          & 0.06                                                                     \\ \hline
    \end{tabular}
    \caption{Performance of different TD algorithms for \emph{Jigsaw Toxicity} dataset}
    \label{tbl:tdAlgorithmsJigsawToxicity}
\end{table*}

\section{H4: \new{Unfair Workers Negatively Impact the Downstream ML Tasks.}}
\label{sec:h4}

Machine learning algorithms often rely on large amounts of data to learn patterns and make predictions. 
However, using data collected from unreliable or biased sources can have severe implications for downstream ML tasks.
% it can lead to severe consequences for downstream ML tasks. 
One potential source of biased data is crowd workers, who may have inherent biases or personal beliefs that impact the quality of the data they provide. 
% One such source of biased data is the use of crowd workers, 
% who may have their own inherent biases that can affect the quality of the data they provide. 
In this scenario, it is crucial to understand 
the impact of unfair crowd workers on downstream machine learning tasks. 
This is of utmost importance as these biases can perpetuate 
and amplify existing societal biases, 
leading to discrimination and unequal treatment of certain groups.

In this section, we study this impact for two representative classifiers -- 
Logistic Regression and Random Forest.
Similar results are observed for other classifiers, such as Support Vector Machines or Neural Networks.
Our experiment is conducted as follows.
We train an ML model $\mathcal{M}_G$, using the ground truth labels.
This is the ideal model we would have obtained if the TD algorithms had been perfect.
Note that the real ground truth is often unavailable in the real world.
Nevertheless, we are able to conduct this experiment due to their availability for our two datasets.
Next, we train a ML model $\mathcal{M}_{TD}$
based on the approximate labels that are produced by the TD algorithm.
We split the dataset equally into training and test sets 
and repeated the experiment 10 times to compute the average accuracy and fairness metrics. 
By comparing the performance of the two models on the same test set,
we can quantify the penalty paid by the TD algorithm.

Tables \ref{tab:4_cj} and \ref{tab:h4_jt}
presents the result for the two datasets, two classifiers, and three TD algorithms.
To conserve space, we only report the delta between the metrics.
Specifically, if the accuracy of the two models is $a_G$ and $a_{TD}$ respectively, 
we report $a_G - a_{TD}$.
We follow the same procedure for the fairness metrics.
As shown in the tables, using TD algorithms results in less accurate and more unfair models. 
This is not surprising as the TD algorithms are not specifically designed to optimize for fairness.

\begin{table*}[!htb]
\scriptsize
    \begin{tabular}{|l|l|c|c|c|}
    \hline
    \textbf{Classifier} & \textbf{TD Algorithm} & \textbf{Delta Accuracy} & \textbf{Delta Demo Par. Diff} & \textbf{Delta Eq. Odds. Diff} \\  \hline
    Random Forest       & Majority Voting       & 6.04                    & 0.06                          & 0.07    \\
    Random Forest       & Dawid-Skene           & 7.63                    & 0.04                          & 0.06     \\
    Random Forest       & Learning from Crowd   & 7.54                    & 0.04                          & 0.05      \\
    Logistic Regression & Majority Voting       & 2.55                    & 0.04                          & 0.07     \\
    Logistic Regression & Dawid-Skene           & 2.77                    & 0.05                          & 0.08      \\
    Logistic Regression & Learning from Crowd   & 2.86                    & 0.05                          & 0.07       \\ \hline
    \end{tabular}
    \caption{Performance of ML model trained with the output of fairness unaware TD algorithm for \emph{Crowd Judgement}}
    \label{tab:4_cj}
\end{table*}

\begin{table*}[!htb]
\scriptsize
    \begin{tabular}{|l|l|c|c|c|}
    \hline
    \textbf{Classifier} & \textbf{TD Algorithm} & \textbf{Delta Accuracy} & \textbf{Delta Demo Par. Diff} & \textbf{Delta Eq. Odds. Diff} \\  \hline
    Random Forest       & Majority Voting       & 5.44                                         & 0.07                                               & 0.09                                               \\
    Random Forest       & Dawid-Skene           & 4.63                                         & 0.05                                               & 0.07   
    \\
    Random Forest       & Learning from Crowd   & 6.23                                         & 0.05                                               & 0.06  \\        
    Logistic Regression & Majority Voting       & 4.32                                         & 0.05                                               & 0.01    \\       
    Logistic Regression & Dawid-Skene           & 3.99                                         & 0.05                                               & 0.08    \\        
    Logistic Regression & Learning from Crowd   & 3.96                                         & 0.04                                               & 0.07      \\   \hline   
    \end{tabular}
    \caption{Performance of ML model trained with the output of fairness unaware TD algorithm for \emph{Jigsaw Toxicity}}
    \label{tab:h4_jt}
\end{table*}

% \subsection{H6: Clean Lab}

\section{H5: \new{The Complementarity of Fair ML and Fair TD.}} 
\label{sec:h5}

% \red{Sara: I only wrote the first section and let the reminder as is.}

% \simone{I updated the plots, I think some of the conclusions we draw should be changed to reflect it}

% From the experiments reported in the previous sections,
Based on the results of the previous experiments,
it is clear that major TD algorithms perform sub-optimally in the presence of unfair workers.
One might wonder whether the extensive amount of prior work on fair ML can mitigate the impact of unfair labels produced by TD algorithms.
% can overcome the unfair labels outputted by TD algorithms.
In other words, can the fair ML algorithms
that seek to produce outputs that are fair to different subgroups 
overcome incorrect/unfair labels produced by TD algorithms?
This section aims to provide a comprehensive comparison of the two possible approaches, fair ML and fair TD, % \hadis{I think two refers to the fairTD which needs to be added here}
and gain insights into the strengths and limitations of each method in 
achieving fairness in decision-making systems.

\begin{figure}[!htb]
\begin{subfigure}{0.49\linewidth}
\includegraphics[width=\textwidth]{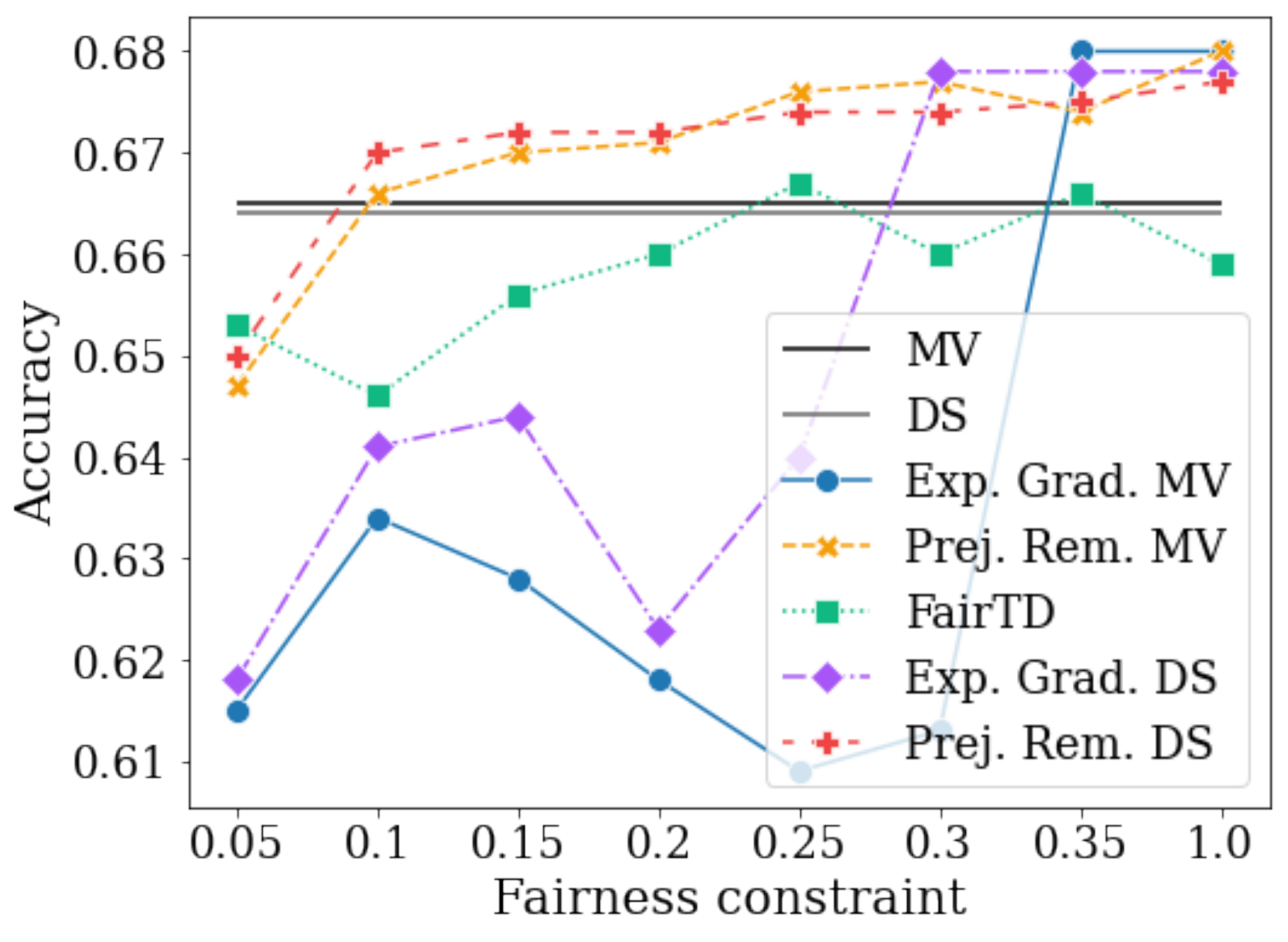}
\caption{Fair ML vs Fair TD for \emph{Crowd Judgement}}
\end{subfigure}
\hfill
\begin{subfigure}{0.49\linewidth}
\includegraphics[width=\textwidth]{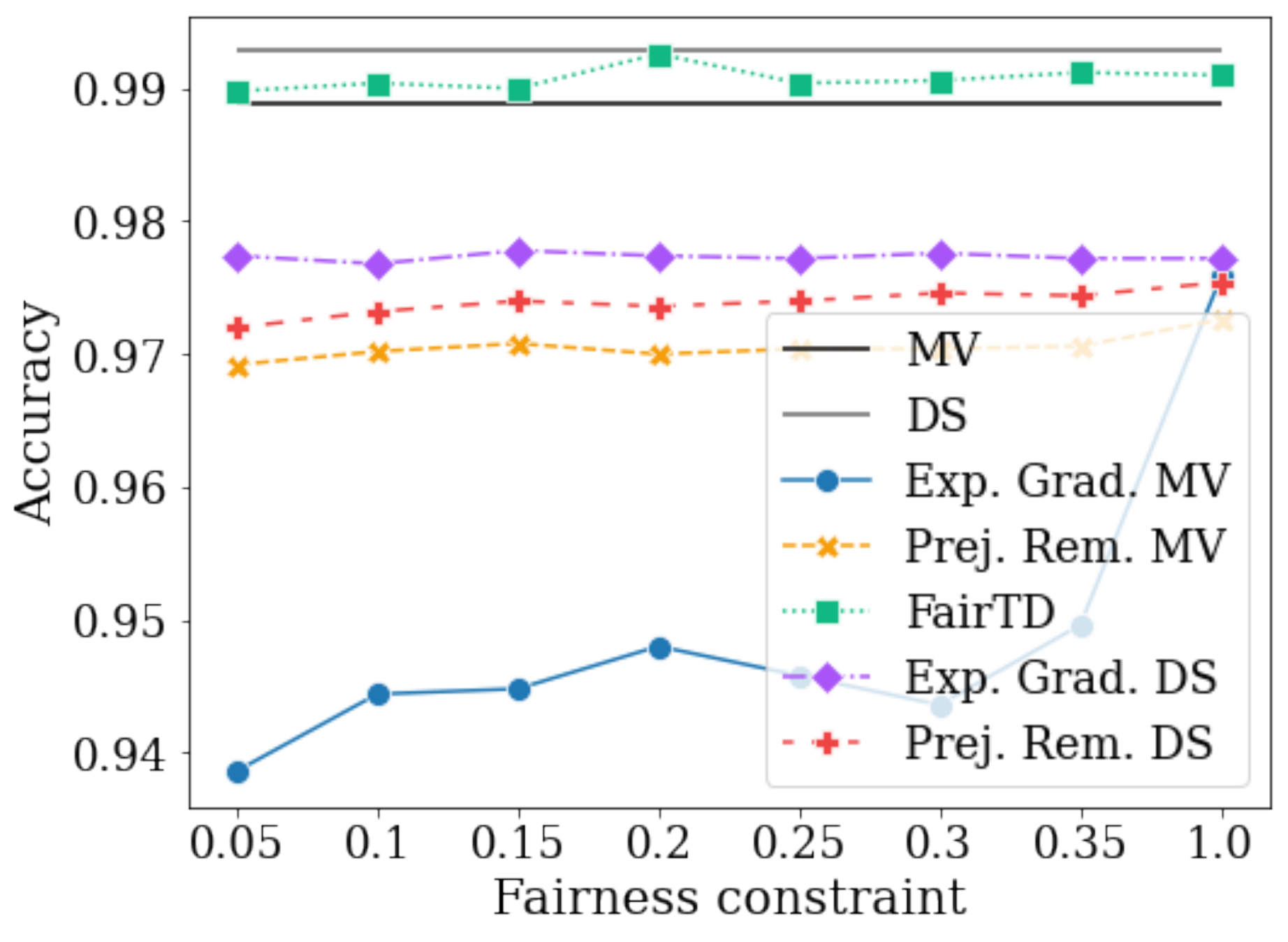}
\caption{Fair ML vs Fair TD for \emph{Jigsaw toxicity}}
\end{subfigure}
\caption{Fair ML vs Fair TD}
\label{fig:h1twitter}
\end{figure}

We compare the output of FairTD, the first and only fairness-aware truth discovery algorithm \cite{kdd2020}, with two fair ML algorithms: Exponentiated Gradient Reduction \cite{agarwal2018reductions} and Prejudice Remover \cite{kamishima2012fairness}. % \hadis{cite each} \simone{I think it's these ones}
% We investigate this by comparing the output of FairTD, 
% the in-processing algorithm proposed by \simone{Li, Yanying
% , et al.} \cite{kdd2020}.
% \simone{To the best of our knowledge, }This is the first and only truth discovery algorithm that is fairness aware.
% We compare its output against two fair ML algorithms -- 
% Exponentiated Gradient Reduction and Prejudice Remover.
We utilize these approaches for both the Majority Voting (MV) and Dawid-Skene (DS) truth discovery strategies.
The experiment's results are presented in Figure \ref{fig:h1twitter}, where the accuracy of the considered approaches is compared with changes in the fairness constraint. The fairness constraint pertains to the allowable violation of fairness for FairTD and Exponentiated Gradient methods, and the reciprocal of the regularization term for Prejudice Remover. While increasing the fairness constraint violation led to an increase in accuracy in some cases, this relationship is not consistent across all datasets and methods tested. Furthermore,  this observation may not hold true when fair workers are consistently more accurate than unfair ones.
We can observe that Prejudice Remover generally outperforms Exponentiated Gradient in mitigating bias while maintaining high accuracy. FairTD, on the other hand, appears to be a robust method that provided good results in both cases, often matching or outperforming Prejudice Remover even for strict constraint values.

Based on our results, it appears that utilizing FairTD, either alone or in conjunction with Fair ML, might be a more promising approach for future research compared to relying solely on fair ML. Although Exponentiated Gradient and Prejudice Removal are valuable techniques in their own right, FairTD provides a robust and precise approach to reducing bias. Therefore, we believe that future research should concentrate on enhancing and optimizing the FairTD method to make it even more effective in addressing the challenges of TD algorithmic fairness.

% Our results suggest that FairTD, or the complementarity of FairTD and Fair ML, might be better options 
% for future studies compared to fair ML alone. 
% While Exponentiated Gradient and Prejudice Removal are useful methods in their own right, 
% FairTD also offers a robust and accurate approach to mitigating bias in algorithmic systems. 
% We believe that future studies should focus on further developing 
% and refining the FairTD method, 
% in order to make it even more effective in addressing the challenges of algorithmic fairness.

%\begin{figure}[!htb]
%\begin{subcaptionblock}{0.49\linewidth}
%\includegraphics[width=\textwidth]{plots/h5_cj.pdf}
%\caption{Fair ML vs Fair TD for \emph{Crowd Judgement}}
%\end{subcaptionblock}
%\hfill
%\begin{subcaptionblock}{0.49\linewidth}
%\includegraphics[width=\textwidth]{plots/h5_jt.pdf}
%\caption{Fair ML vs Fair TD for \emph{Jigsaw toxicity}}
%\end{subcaptionblock}
%\caption{Fair ML vs Fair TD}
%\label{fig:h1twitter}
%\end{figure}

\section{Conclusion}
\label{sec:conclusion}

In this paper, we investigated a key source of bias in ML datasets produced using crowdsourcing.
Most of these datasets use simple majority voting to aggregate conflicting responses of crowd workers.
However, this approach is sub-optimal and produces poor results in the presence of unfair workers.
We conducted extensive experiments to study the impact of unfair workers on various truth discovery algorithms.
and conclude that their impact on label accuracy is significant.
Existing truth discovery algorithms are focused on accuracy and often cannot handle unfair workers.
It is important to design fairness-aware truth discovery algorithms 
so that one could counteract the influence of unfair workers on the labels for ML datasets. 

\balance
\bibliographystyle{siam}
\bibliography{ref}

\end{document}